\documentclass[letterpaper, 10 pt, conference]{ieeeconf}  % Comment this line out if you need a4paper

\IEEEoverridecommandlockouts                              % This command is only needed if 
\usepackage{amsmath,amsfonts}
\usepackage{algorithmic}
\usepackage{array}
\usepackage{textcomp}
\usepackage{stfloats}
\usepackage{url}
\usepackage{graphicx}
\usepackage{booktabs} 
\usepackage{pifont}
\usepackage{hyperref}       % hyperlinks

\newcommand{\methodname}{CLAR}

\begin{document}
% \title{CL3R: 3D Reconstruction and Contrastive Learning for Enhanced Robotic Manipulation Representations}
% \title{CLAR: A Robotic Representation Learning Framework Integrating Contrastive Learning, Local Alignment, and 3D Reconstruction}

\title{\LARGE \bf
CLAR: Learning 3D Representations for Robotic Manipulation by Fusing Masked Reconstruction with Multi-Level Contrastive Alignment
}

% \title{CLAR: Learning 3D Representations for Robotic Manipulation by Fusing 3D Masked Reconstruction with Multi-Level Contrastive Alignment}

% \title{CLAR: Fusing 3D Masked Reconstruction with Multi-Level Contrastive Alignment to Learn Representations for Robotic Manipulation}

\author{
  Wenbo Cui\textsuperscript{* 1,2,4}, 
  Chengyang Zhao\textsuperscript{* 3}, 
  Yuhui Chen\textsuperscript{1,2}, 
  Haoran Li\textsuperscript{1,2}, \\
  Zhizheng Zhang\textsuperscript{4,5}, 
  Dongbin Zhao\textsuperscript{1,2}
  He Wang\textsuperscript{\dag \ 4,5}\\
\thanks{* Equal Contribution.}
\thanks{\dag \ Corresponding author}
\thanks{This work was supported by the Suzhou Innovation and Entrepreneurship Leading Talents Programme - Innovation Leading Talent in Universities and Research Institutes with Grant No. ZXL2025310}
% \thanks{This work is supported by the Excellent Youth Program of State Key Laboratory of Multimodal Artificial Intelligence Systems.}
\thanks{$^{1}$SKL-MAIS, Institute of Automation, Chinese Academy of Sciences,
        $^{2}$School of Artificial Intelligence, University of Chinese Academy of Sciences,
        % $^{3}$Beijing Academy of Artificial Intelligence, Beijing, China,
        $^{3}$Carnegie Mellon University,
        $^{4}$Galbot,
        $^{5}$CFCS, School of Computer Science, Peking University,
}
}

% \author{Anonymous
% % \thanks{Manuscript created October, 2020; This work was developed by the IEEE Publication Technology Department. This work is distributed under the \LaTeX \ Project Public License (LPPL) ( http://www.latex-project.org/ ) version 1.3. A copy of the LPPL, version 1.3, is included in the base \LaTeX \ documentation of all distributions of \LaTeX \ released 2003/12/01 or later. The opinions expressed here are entirely that of the author. No warranty is expressed or implied. User assumes all risk.}
% }

\maketitle
\thispagestyle{empty}
\pagestyle{empty}

The spatial information inherent in 3D point clouds is crucial for robotic manipulation. However, existing 3D pre-training methods face a fundamental trade-off: Masked Autoencoding (MAE) excels at capturing spatial-geometric features but lacks semantics, whereas contrastive learning, while able to distill semantics from 2D foundation models, is ill-suited for the fine-grained details required for manipulation tasks.
To address these challenges, we propose \textbf{\textit{\methodname}}, a novel 3D pre-training framework that synergizes global understanding with fine-grained local alignment. Our framework unifies MAE with global cross-modal contrastive learning to integrate robust spatial awareness with rich semantic understanding. To enhance its focus on fine-grained details, at the local level, we introduce an adaptive alignment mechanism that leverages deformable attention to force precise correspondences between local 3D geometry and 2D visual features, thereby overcoming the limitations of conventional global alignment in manipulation tasks.
Extensive experiments in simulation and the real world demonstrate that \textit{\methodname} achieves state-of-the-art performance, significantly outperforming existing methods in visuomotor policy learning. 
Our project page is 
\href{https://cwb0106.github.io/CLAR/}{https://cwb0106.github.io/CLAR/}.

\section{INTRODUCTION}

Building a robust representation module is essential for visuomotor policy learning, enabling robotic systems to perform various everyday tasks~\cite{chen2025conrft, robogpt2023, li2024generalizing}. Many existing studies~\cite{lift3d2024, goyal2023rvt, goyal2024rvt2learningprecisemanipulation} highlight the advantages of integrating pre-trained 2D vision pre-training models into the perception module of robotic policies, leveraging them as powerful vision encoders. This approach allows policies to inherit the strong semantic understanding capabilities of foundation models while facilitating rapid adaptation to robotic tasks through fine-tuning on task-specific datasets.

\begin{figure*}[ht]
% \captionsetup{skip=5pt}
    \centering
    \includegraphics[width=0.95\textwidth, ]{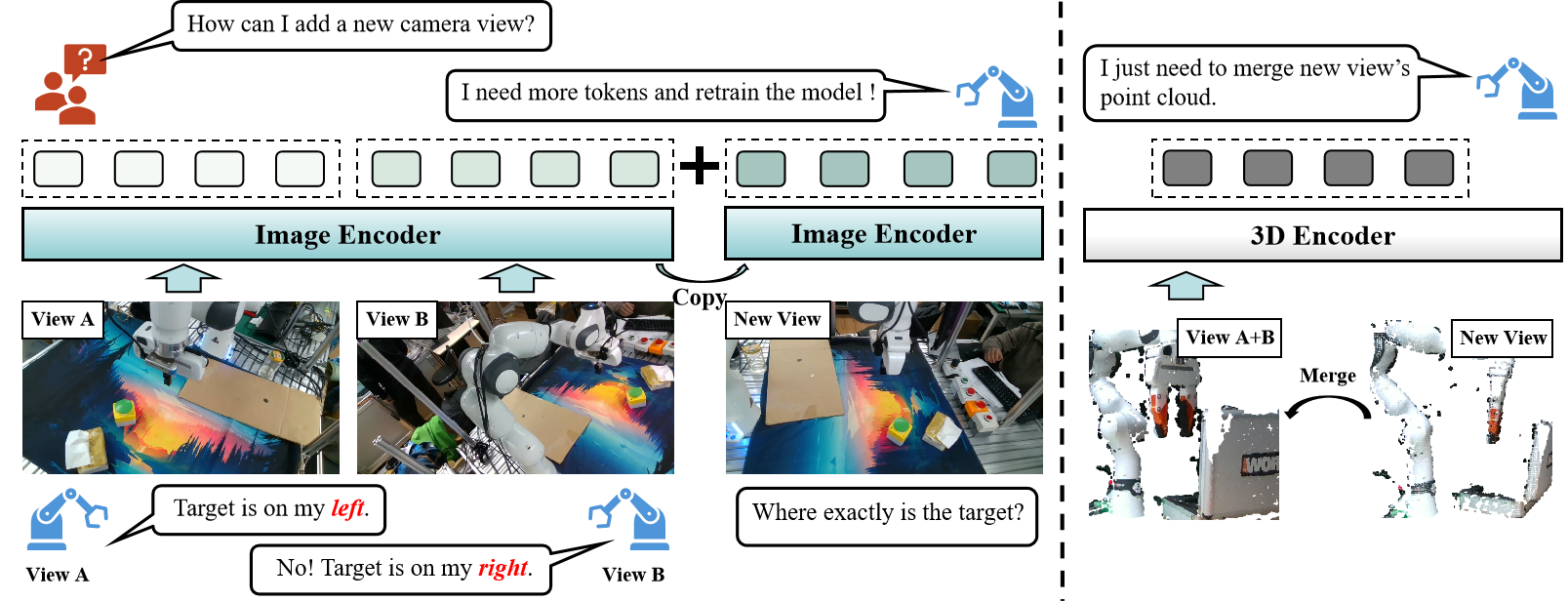}
    \vspace{-3pt}
% \caption{\textbf{The distinction between 2D and 3D modalities in robotic pre-training}. When a robot requires additional camera perspectives due to a limited field of view, 2D-based methods necessitate encoding more tokens and fine-tuning or pre-training the model. In contrast, 3D-based methods can efficiently enhance visual input through point cloud fusion. Additionally, as shown in the bottom-left diagram, the absence of a unified spatial coordinate system in 2D-based methods can cause multi-view ambiguity when observing the same objects from different perspectives, hindering spatial awareness.}
\caption{\textbf{A Comparison of 2D and 3D Modalities for Robotic Pre-training.} 
\textbf{Left:} 2D-based methods lack a unified spatial coordinate system, leading to multi-view ambiguity where an object's relative position (e.g., "left" vs. "right") is inconsistent across views. This forces the policy to learn complex, view-dependent mappings, making it difficult to incorporate new camera views without costly retraining. 
\textbf{Right:} 3D-based methods using point clouds operate in a unified, robot-centric coordinate frame. This resolves view ambiguity and allows new visual information to be seamlessly integrated via simple geometric fusion, enhancing the policy's adaptability and spatial awareness.}
    \label{fig:data_modalities}
    \vspace{-15pt}
\end{figure*}

Despite its benefits, 2D visual pre-training models still face fundamental bottlenecks in robotic manipulation.
First, they lack 3D spatial awareness~\cite{lift3d2024}. Their 2D backbones, engineered for planar images, are inherently ill-equipped to understand scene geometry, which prevents policies from effectively grounding themselves in the physical 3D world. Incorporating multi-view or depth information provides only marginal improvements. Second, the absence of a unified, robot-centric coordinate system creates severe multi-view ambiguity. This forces policies to learn brittle, camera-centric behaviors—for instance, an object appearing on the robot's "left" from one camera and its "right" from another (Fig.~\ref{fig:data_modalities})—rather than forming a coherent understanding of spatial relationships. As a result, these learned skills do not generalize to different robot or camera setups, drastically limiting their autonomy and adaptability for real-world deployment.

To address the limitations of 2D foundation models in robotics tasks, recent works have shifted to using point clouds as observation for policy learning~\cite{rlafford2023, gapartmanip}. While leveraging pre-training to circumvent the poor robustness of training from scratch~\cite{lift3d2024} is a consensus, existing 3D pre-training methods face significant and distinct hurdles.
On one hand, approaches that lift features from 2D models to 3D~\cite{lift3d2024, qu2025spatialvla} are constrained by the modality gap, which can cause information loss~\cite{lift3d2024}. On the other hand, methods for pure 3D pre-training face their own inherent challenges. These methods exhibit a split between Masked Autoencoding (MAE) for spatial geometry~\cite{pointmae2022} and contrastive learning for semantic understanding~\cite{ulip2024}, yet a fusion of their strengths is crucial for robotic manipulation. Moreover, the application of contrastive learning to robotics is flawed: its global features loses local details, and point cloud cropping creates contextual mismatch that degrades representation quality. These combined challenges make it particularly difficult to design a pre-training framework that balances both geometry and semantics for robotics tasks.

In this paper, we introduce \textbf{\textit{\methodname}}, a novel pre-training method that enhances robotic representations by unifying \textbf{C}ontrastive \textbf{L}earning for semantic understanding, local \textbf{A}lignment for fine-grained details, and 3D \textbf{R}econstruction for spatial perception.
 The key insight behind \textit{\methodname} is that both spatial awareness and advanced semantic understanding are equally crucial for developing a powerful and robust representation model in manipulation tasks.
To enhance spatial understanding, we employ a point cloud MAE within an encoder-decoder architecture. This approach enables the model to learn high-level latent features from unmasked point patches while reconstructing masked ones, thereby strengthening its spatial reasoning capabilities~\cite{pointmae2022}.
For robust semantic understanding, we leverage pre-trained 2D foundation models and use a contrastive learning-based mechanism to align the point cloud feature space with the powerful and expressive feature space of foundation models (such as CLIP). This alignment facilitates rapid semantic knowledge transfer from the 2D foundation model to our 3D representation model without requiring extensive 3D pre-training data, thereby mitigating the challenges posed by limited 3D robotic datasets~\cite{ulip2024}.
Furthermore, to enhance perception of scene details and resolve the 'contextual mismatch' common in robotics pre-training, we design a novel local feature alignment strategy. Instead of forcing a flawed global alignment between a cropped point cloud and a full image, our strategy uses deformable attention~\cite{deepinteraction++} to adaptively match local 3D patches with corresponding 2D regions. This achieves a robust cross-modal local alignment that focuses learning on meaningful, shared information, respecting the irregular structure of the point cloud.

To summarize, our main contributions are as follows:

\begin{itemize}

\item We propose \textit{\methodname}, a novel representation learning framework tailored for robotic manipulation. \textit{\methodname} integrates 3D reconstruction to enhance spatial perception and employs global contrastive learning to strengthen semantic understanding.

\item We propose a novel adaptive local alignment mechanism designed to preserve the fine-grained details essential for manipulation, overcoming the limitations of global alignment. By leveraging deformable attention for adaptive local feature matching, our mechanism also effectively resolves the contextual mismatch introduced by standard cropping procedures.

\item We conduct extensive experiments to validate the effectiveness of \textit{\methodname}. It outperforms state-of-the-art (SOTA) approaches in visuomotor robotic manipulation, with results on MetaWorld (82.6\% vs. 76.8\%), RLBench (82.0\% vs. 77.0\%), and real-world tasks (83.0\% vs. 61.0\%), demonstrating enhanced spatial awareness and semantic understanding.

\end{itemize}

\vspace{-5pt}

\section{RELATED WORK }

\subsection{3D Representation Pre-training}

Recent studies attempt to enhance the 3D understanding capabilities by pre-training on 3D point clouds~\cite{pointmae2022, ulip2024, uni3d2024, recon2023, shapellm2024}. PointMAE~\cite{pointmae2022} and PointBert~\cite{pointbert2022} leverage MAE to reconstruct masked point clouds for strengthening the models' geometric comprehension. Following the success of contrastive learning in 2D foundation models, \cite{ulip2024, uni3d2024} apply contrastive learning for aligning textual, image, and point cloud features to facilitate the learning of 3D representations. Recon~\cite{recon2023} and ShapeLLM~\cite{shapellm2024} attempt to combine both methods simultaneously, achieving promising results. However, these approaches have been trained and tested on datasets comprising virtually limitless object data, making them challenging to apply to robotic tasks characterized by data scarcity, complex environments, and significant variations in point cloud scales. Crucially, these methods typically perform feature alignment at a global level, overlooking the fine-grained, local correspondences that are essential for precise manipulation tasks. As a result, these pre-trained models encounter difficulties when applied to robotic tasks. Our aim is to propose a 3D representation learning framework tailored for robotic manipulation tasks, capable of acquiring spatial comprehension skills while also attaining a degree of advanced semantic understanding.

\subsection{Robot Representation Learning}

With the maturation of the pre-trained visual representation learning paradigm, many studies~\cite{lift3d2024, r3m2023, spa2024, vc12023} focus on enhancing visual representations in robotics using pre-training models. Currently, most efforts are concentrated on 2D-based representations, with frameworks such as OpenVLA~\cite{openvla2024}, and RDT~\cite{rdt2024} utilizing pre-trained 2D foundation models for extracting visual features. R3M \cite{r3m2023} attempts to learn visual representations from human video data through contrastive learning, while MVP~\cite{mvp2023}, VC-1~\cite{vc12023}, and Voltron~\cite{voltron2023} apply MAE to strengthen the visual representation capabilities in robotics. However, these 2D-based methods lack spatial awareness and suffer from challenges in cross-dataset camera view ambiguities, making it difficult to obtain a robust robotic representation module through pre-training on effective robotic datasets. Other studies also attempt to leverage 2D modality data to learn 3D representation capabilities. Lift3D~\cite{lift3d2024}, MV-MWM~\cite{MV-MWM2023}, 3D-MVP~\cite{3dmvp2024}, and SPA~\cite{spa2024} aim to acquire 3D visual representations through multi-view data and depth image. While these methods strengthen the spatial awareness of 2D models, they remain imprecise in capturing fine geometric structures due to limitations of data modality. Unlike these methods, our method~\textit{\methodname} conducts 3D modality pre-training, leveraging multi-view embodied datasets to enhance spatial awareness and semantic understanding capabilities in robotic manipulation tasks.

\begin{figure*}[ht]
% \captionsetup{skip=5pt}
    \centering
    \includegraphics[width=0.95\textwidth, ]{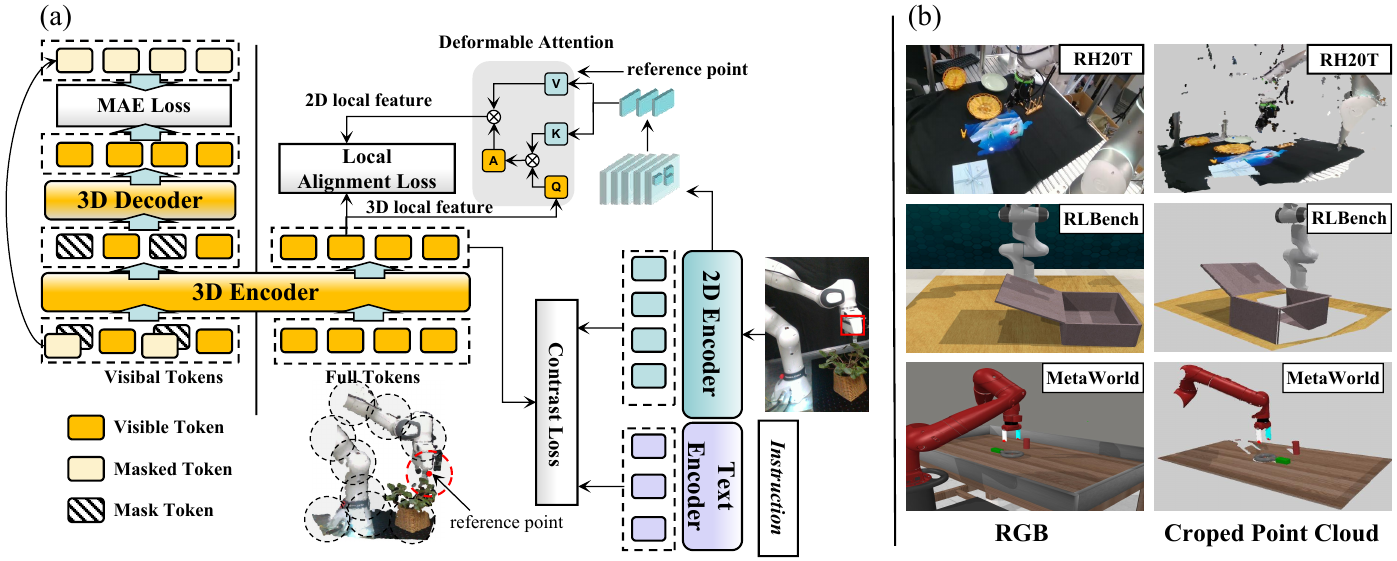}
    \vspace{-3pt}
\caption{\textbf{(a) The \textit{\methodname} Pre-training Framework}. \textbf{\textit{\methodname}} enhances spatial understanding via MAE and semantic comprehension through contrastive learning. To capture the fine-grained local details essential for robotic tasks, we supplement the global contrastive loss with an adaptive local feature alignment mechanism.  \textbf{ (b) Contextual Mismatch Induced by Point Cloud Cropping}. The common practice of cropping a point cloud in robotics removes background context, leading to a feature discrepancy between the partial point cloud and the full RGB image.}
    \label{fig:pipeline}
    \vspace{-15pt}
\end{figure*}

\section{METHOD}

\subsection{Globla Contrastive Learning and 3D Reconstruction for Robotic Representation}

% Existing 3D pre-training approaches exhibit three critical limitations in developing representation models with robust spatial and semantic comprehension for robotic manipulation tasks. Primarily, most methods are trained on conventional object-centric 3D datasets, resulting in features that are not directly transferable to more complex scene-level robotic applications. Secondly, these approaches inadequately integrate the geometric reconstruction advantages of masked autoencoding (MAE) with the semantic distillation capabilities of contrastive learning from 2D visual models. Ultimately, even advanced 3D contrastive learning frameworks predominantly focus on global feature alignment, overlooking the essential fine-grained local representations required for precise robotic manipulation, consequently limiting their effectiveness in complex interactive operations.

Existing 3D pre-training approaches often struggle to generalize from conventional object-centric datasets to complex robotic scenes, and inadequately integrate the geometric reconstruction advantages of masked autoencoding (MAE) with the semantic distillation capabilities of contrastive learning. 
To address these limitations, we employ 3D point clouds as input to enhance spatial representation and enforce coordinate consistency in 3D pre-training models. Our key insight is that both spatial awareness and semantic understanding are essential for robotic tasks. To enhance spatial awareness, we utilize point cloud MAE in \textit{\methodname}, while cross-modal contrastive learning is employed to effectively transfer semantic knowledge from pre-trained 2D foundation models into \textit{\methodname}, as illustrated in Fig.~\ref{fig:pipeline}-(a). First, during pre-training, we project all 3D point cloud data into a unified robot coordinate system using the extrinsic parameters of each camera. This step mitigates ambiguity caused by varying camera viewpoints across datasets, enabling our method to establish generalizable 3D perception with fewer 3D data during pre-training. Next, given an input point cloud  $X \in \mathbb{R}^{N \times 3}$ with $N$ points, we use the farthest point sampling (FPS) to sample $n$ center points $P_c \in \mathbb{R}^{n \times 3}$. We then apply the K-Nearest Neighbors (KNN) algorithm to search the neighborhoods of these $n$ points, generating corresponding $n$ point patches $P \in \mathbb{R}^{n \times k \times 3}$. A high masking ratio $m$ (60\%-80\%) is randomly applied to $P$, with the masked patches denoted as $P_{gt} \in \mathbb{R}^{mn \times k \times 3}$, which is used as the ground truth (GT) for computing the reconstruction loss. The visible point patches $P_v \in \mathbb{R}^{(1-m)n \times k \times 3}$ and full point patches $P$ are embedded into visible tokens $T_v$ and full tokens $T$, respectively. Subsequently, a standard transformer $f_\theta(\cdot)$ embeds $T_v$  and $T$, along with their positional embeddings, generating encoded visible patches $\mathcal{F}_v^P$ and encoded complete patches $\mathcal{F}^P$, respectively.

\textbf{Enhancing Geometric Perception with MAE.} 
\textit{\methodname} inputs $\mathcal{F}_v^{P}$ and masked tokens $\mathcal{F}_m^P$ into the MAE decoder, where a complete set of positional embeddings is added to each transformer block. The final reconstructed point patches $P_{pre}$ are obtained after passing through a prediction head. The MAE loss is computed as the $\ell 2$ Chamfer Distance between $P_{pre}$ and $P_{gt}$. 

\vspace{-10pt}

\begin{equation}
% \vspace{-12pt}
\small
    \mathcal{L}^{R} = \frac{1}{\left | P_{pre}\right |} \displaystyle\sum_{r \in P_{pre}} \min_{g \in P_{gt}} {\left \| r-g\right \|_2^2} + \frac{1}{\left | P_{gt}\right |} \displaystyle\sum_{g \in P_{gt}} \min_{r \in P_{pre}} {\left \| r-g\right \|_2^2}
% \vspace{-3pt}
\end{equation}

\textbf{Enhancing semantic understanding with contrastive learning.} \textit{\methodname} utilizes pre-trained image encoder $E_I$ and text encoder $E_T$ to extract image features $\mathcal{F}^I$ and text features $\mathcal{F}^T$, respectively. We align  $\mathcal{F}^P$ with $\mathcal{F}^I$ and $\mathcal{F}^T$ through a distillation-like paradigm, which efficiently transfers the semantic understanding capabilities of the pre-trained models to \textit{\methodname}.  The 3D-to-image and 3D-to-text alignment is formulated using a contrastive loss function, which corresponds to the "Contrast Loss" module in Fig.~\ref{fig:pipeline}-(a):

\vspace{-10pt}
% \scalebox{0.7}{
\begin{equation}
% \small
\vspace{-5pt}
\mathcal{L}^{C}_{PI} = -\frac{1}{2}\displaystyle\sum_{i}\left [ 
\mathcal{S}\left ( \mathcal{F}^P, \mathcal{F}^I\right )_i
+
\mathcal{S}\left ( \mathcal{F}^I, \mathcal{F}^P\right )_i
\right ],
\vspace{-5pt}
\end{equation}
% }

\begin{equation}
% \small
\vspace{-5pt}
\mathcal{L}^{C}_{PT} = -\frac{1}{2}\displaystyle\sum_{i}\left [ 
\mathcal{S}\left ( \mathcal{F}^P, \mathcal{F}^T\right )_i
+
\mathcal{S}\left ( \mathcal{F}^T, \mathcal{F}^P\right )_i
\right ],
\vspace{-5pt}
\end{equation}
% }

\begin{equation}
% \small
\vspace{-5pt}
\mathcal{S}\left ( \mathcal{U}, \mathcal{V}\right )_i=
\log \frac{\exp\left (\mathcal{U}_i \mathcal{V}_i/ \tau    \right )}{\displaystyle\sum_{j, i \ne j}\exp\left ( \mathcal{U}_i \mathcal{V}_j / \tau \right )},
% \vspace{-5pt}
\end{equation}

\noindent where $\tau$ is a learnable temperature parameter, and $i$, $j$ are the sampling indices.

\subsection{Adaptive Fine-Grained Cross-Modal Alignment of Local Features.}

To overcome the failure of global feature alignment in capturing fine-grained details for manipulation, we introduce an adaptive local alignment mechanism using deformable attention. This forces precise local 3D-to-2D correspondences, capturing the essential details for manipulation while simultaneously resolving the contextual mismatch caused by cropping techniques, as visually illustrated in Fig.~\ref{fig:pipeline}-b.
Specifically, our mechanism performs this alignment via cross-attention.
The process begins by constructing a dense mapping $M^{P->I}$ from the point cloud coordinate system $P$ to the image coordinate system $I$, which projects a 3D spatial point $(x^P, y^P, z^P)$ to a 2D pixel coordinate $(u^I, w^I)$.
For each 3D point cloud patch, its encoded geometric feature serves as the \textbf{query} $q=\mathcal{F}_{\text{local}}^{P}$. The entire 2D image feature map serves as the \textbf{value} $\mathbf{v}$. However, since 3D point cloud patches are defined by KNN, they have a variable spatial scale. This presents a mismatch with the typically fixed receptive fields used on 2D images. 

To resolve this mismatch, we employ deformable attention~\cite{deepinteraction++, deformable2020} to extract local features. Departing from conventional methods that attend to features within a static, grid-like receptive field, the core of our approach is using the local patch centers of the point cloud as queries. Since the queries originate from the 3D point cloud, the attention mechanism is guided to sample features only from corresponding valid regions within the 2D image, thereby overcoming the contextual mismatch caused by point cloud cropping. For each query $q$ with a corresponding 2D reference point $p_q$, the model predicts sampling offsets $\Delta p_{qk}$ and attention weights $A_{qk}$. The aligned 2D visual feature is then computed as a weighted sum of the sampled features:

\vspace{-5pt}

\begin{equation}
\mathcal{F}_{\text{local}}^{I} = \sum_{k=1}^{K} A_{qk} \cdot \mathbf{v}(p_q + \Delta p_{qk})
\vspace{-3pt}
\end{equation}

\noindent where $p_q$ is the normalized 2D coordinate projected from the 3D patch center, while $\Delta p_{qk}$ and $A_{qk}$ are learned adaptively from the query $q$. This mechanism dynamically adjusts its receptive field on the image according to the input 3D query, allowing it to flexibly capture visual information corresponding to 3D geometric structures of varying scales. This strategy ensures that the model learns precise and robust local correspondences between the 3D shape and its 2D appearance. The Local Alignment Loss shown in Fig.~\ref{fig:pipeline}-(a),  denoted as $\mathcal{L}_{PI}^{local}$ :

\vspace{-7pt}

\begin{equation}
\small
\mathcal{L}^{local}_{PI} = -\frac{1}{2}\displaystyle\sum_{i}\left [ 
\mathcal{S}\left ( \mathcal{F}_{local}^P, \mathcal{F}_{local}^I\right )_i
+
\mathcal{S}\left ( \mathcal{F}_{local}^I, \mathcal{F}_{local}^P\right )_i
\right ]
\end{equation}

Finally, the total loss L is a weighted sum of all components:

\vspace{-3pt}

\begin{equation}
\mathcal{L} = \alpha \mathcal{L}^{R} + \beta \mathcal{L}^{C}_{PI} + \gamma \mathcal{L}^{C}_{PT} + \theta\mathcal{L}_{PI}^{local}
\end{equation}

\noindent where $\alpha=1.5$, $\beta=0.5$, $\gamma=0.5$ and $\theta=0.5$ represent the weighting coefficients for each component of the loss function.

\subsection{Pre-training Dataset}
 Due to the scarcity of embodiment data containing camera parameters and depth images, we select RH20T~\cite{rh20t2023} and RLBench~\cite{rlbench2020} datasets for pre-training. RH20T contains 110K contact-rich robot trajectories, over 140 tasks, and a large multi-view dataset with RGB, depth, and other modalities, providing a strong foundation for unified spatial coordinate systems. To compensate for the lack of real-world data, we supplement it with simulation data from RLBench (\textit{close-jar}, \textit{insert-onto-square-peg}, \textit{move-hanger}, \textit{open-drawer}, \textit{push-buttons}, \textit{put-groceries-in-cupboard}, \textit{stack-blocks}, \textit{stack-cups}, \textit{turn-oven-on}). By co-training  RH20T real-world data with RLBench simulation data, we partially alleviate the data scarcity issue.

\begin{figure}[ht]
\vspace{-8pt}
% \captionsetup{skip=5pt}
    \centering
    \includegraphics[width=0.45\textwidth, ]{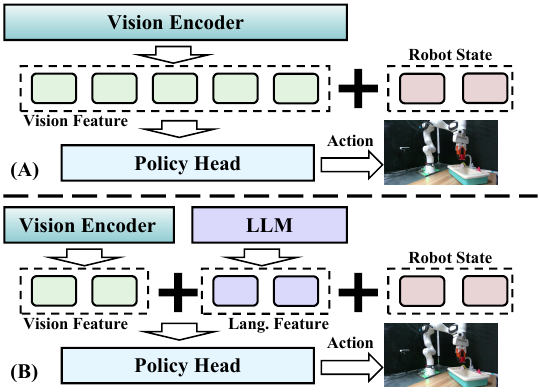}
    % \caption{\textbf{Policy Architecture for Imitation Learning.} Depending on the tested algorithm, the vision encoder can be replaced with the corresponding image encoder or 3D encoder. \textbf{(A) Policy architecture under standard experimental settings} (non-multi-task experiments). The 2D/3D vision pre-training model outputs visual features, which are concatenated with robot state information and jointly fed into the downstream MLP-based policy head to predict actions. \textbf{(B) Policy architecture in multi-task experiments.} Visual features from the vision encoder, language features from the LLM, and robot state information are combined and fed into the policy head. In both experimental settings, vision and language features share the same dimensionality.}
    \caption{\textbf{Policy Architecture for Imitation Learning.} (A) Single-task experimental setup. (B) Multi-task experimental setup.}
    \label{fig:policy}
    \vspace{-15pt}
\end{figure}

% \noindent \textbf{Policy architecture.} 
\subsection{Policy Architecture}

We evaluate all models using a unified imitation learning policy architecture, as depicted in Fig.~\ref{fig:policy}. The policy head is consistently an MLP-based structure.
In standard, single-task experiments (non-multi-task experiments, Fig.~\ref{fig:policy}-A), the input to the policy head is formed by concatenating two components: the visual features from the vision encoder being evaluated and the robot's proprioceptive state.
For multi-task experiments (Fig.~\ref{fig:policy}-B), an additional modality is incorporated into the input: language features representing the task instruction, which are encoded by a pre-trained LLM. In all settings, features from different modalities (such as vision and language) are projected to a shared dimensionality before being fused and processed by the policy head to predict the action.

\section{EXPERIMENT}

% We present the experimental setups in Section 4.1, including both simulation and real-world experiments. Section 4.2 evaluates the overall performance of each method on simulation benchmarks. In Section 4.3, we conduct a perspective ambiguity experiment to verify whether there is a problem of multi-view ambiguity in 2d-based pertaining methods, and whether 3D-based methods can overcome this difficulty. In Section 4.4, we present multitask experiments to validate the semantic understanding capabilities of each method. 
% The ablation experiment and real world experiment are presented in Sections 4.4 and 4.5, respectively.

\begin{table*}[ht]
\centering
\caption{
Quantitative Results (Success Rate, \%) on Representation Models (MetaWorld \& RLBench)
% 2D Rep. and 3D Rep. denote 2D and 3D representation methods, respectively.
}
\begin{center}
\setlength{\tabcolsep}{3pt} % Adjust the space between columns
\renewcommand{\arraystretch}{1} % Adjust the row height (1 is the default)
\begin{tabular}{l|c|c|c|cccc|cc }
\toprule
Methods & Type  & Input Data  & Pretrained  & Easy(5)  & Medium(5) & Hard(5)  &  Mean & RLBench(4)\\
\midrule

% SGM~\cite{hirschmuller2008sgm} & 5.03 & 1.411 & 0.564 & 0.821 & 22.83 & 25.97 & 29.16 \\
% \midrule
CLIP \cite{clip2021} & 2D Rep. & RGB & \ding{51} & 58.0\% & 62.4\% & 48.8\% & 56.4\% & 48.0\% \\
% \midrule
R3M \cite{r3m2023} & 2D Rep. & RGB & \ding{51} & 80.0\% & 68.8\% & 56.4\% & 68.4\% & 53.0\%\\
VC-1 \cite{vc12023} & 2D Rep. & RGB & \ding{51} & 50.8\% & 72.4\% & 47.2\% & 56.8\% & 43.0\%\\
VGGT \cite{vggt2025}  & 2D Rep. & RGB & \ding{51} & 16.8\% & 17.6\% & 2.4\% & 12.2\% & -\\
\midrule
PointNet \cite{pointnet2017} & 3D Rep. & PC & \ding{56} & 70.4\% & 52.4\% & 44.0\% & 55.6\% & 55.5\%\\
PointNet++ \cite{pointnet++2017} & 3D Rep. & PC & \ding{56} & 82.0\% & 60.8\% & 57.6\% & 66.8\% & 60.0\%\\
PointNext \cite{pointnext2022} & 3D Rep. & PC & \ding{56} & 76.8\% & 63.6\% & 39.2\% & 59.8\% & 56.0\%\\
DP3 \cite{dp32024} & 3D Rep. & PC & \ding{56} & 79.2\% & 51.2\% & 50.5\% & 60.3\% & -\\ 
SPA \cite{spa2024} & 3D Rep. & RGB & \ding{51} & 65.2\% & 76.8\% & 57.6\% & 66.5\% & 63.0\%\\
Lift3D \cite{lift3d2024} & 3D Rep. & PC & \ding{51} & 89.2\% & 80.4\% & 60.8\% & 76.8\% & 77.0\%\\
% RS~\cite{lipson2021raftstereo} & 4.94 & 1.375 & 0.529 & 0.826 & 23.06 & 25.18 & 31.15 \\
\midrule
% DP3 \cite{dp32024} & 3D Policy & PC & \ding{56} & 79.2 & 51.2 & 50.4 & 60.26 & -\\
% \midrule
\textbf{CLAR (Ours)} & 3D Rep. & PC & \ding{51} & \textbf{96.0\%} & \textbf{80.8\%} & \textbf{71.2\%} & \textbf{82.6\%} & \textbf{82.0\%}\\

\bottomrule
\end{tabular}
\end{center}
% \vspace{-15pt}
\label{tab:main_experiment}
\vspace{-3pt}
* 2D Rep. and 3D Rep. denote 2D and 3D representation methods, respectively.
\vspace{-7pt}
\end{table*}

\subsection{Experimental Setup}

 \textbf{Benchmarks.} 
We evaluate various methods on two widely used manipulation benchmarks: MetaWorld~\cite{metaworld2020}, based on MuJoCo, and RLBench~\cite{rlbench2020}, based on CoppeliaSim. We select 15 tasks of varying difficulty levels from MetaWorld: \textbf{easy} (\textit{reach, lever-pull, handle-pull, peg-unplug-side, dial-turn}), \textbf{medium} (\textit{hammer, sweep-into, bin-picking, push-wall, box-close}), and \textbf{hard} (\textit{assembly, hand-insert, shelf-place, pick-place-wall, disassemble}). In RLBench, we select four tasks where the front camera view is employed:  \textit{close-box, close-laptop-lid, water-plants, and toilet-seat-down}.

% \noindent \textbf{Policy architecture.} 
% We evaluate all models using a unified imitation learning policy architecture. For standard, single-task experiments (non-multi-task experiments), the input to an MLP-based policy head consists of features from the vision encoder (the model being evaluated) concatenated with the robot's proprioceptive state. For multi-task experiments, language features from a pre-trained LLM, representing the task instruction, are also concatenated. All feature modalities are projected to a shared dimensionality before being fused and processed by the policy head to predict the action.

 \textbf{Data preparation.} 
Expert policies from MetaWorld and RLBench are used to collect trajectories. For MetaWorld, 30 demonstrations are gathered per task, with 25 used for training and 5 for validation, each containing 200 steps. For RLBench, the demonstrations comprise keyframe waypoint data rather than the dense trajectory data used in MetaWorld.

\textbf{Baselines.} We compare eleven methods across four categories: (1) \textbf{2D pre-training representation methods}, including CLIP~\cite{clip2021} (pre-trained on general internet datasets), VGGT\footnote{For VGGT, we extract the intermediate feature map from its DPT depth head and use a convolutional projection head to aggregate it into a single feature vector for the subsequent action head.}~\cite{vggt2025} (pre-trained on large-scale datasets with 3D supervision) and two robotic-specific methods, R3M~\cite{r3m2023} and VC-1~\cite{vc12023} (both pre-trained on robotic datasets). 
% As for VGGT, we extract the intermediate feature map from its DPT depth head and use a convolutional projection head to aggregate it into a single feature vector for the subsequent action head. 
(2) \textbf{3D representation methods without pre-training}, including widely used approaches such as PointNet~\cite{pointnet2017}, PointNet++~\cite{pointnet++2017}, and PointNext~\cite{pointnext2022}. (3) \textbf{3D robotic pre-training representation methods}, including SPA~\cite{spa2024} and Lift3D~\cite{lift3d2024}. SPA employs differentiable neural rendering on multi-view images to improve 3D representation. Lift3D, a previous SOTA 3D robotic pre-training representation method, employs task-aware MAE during the pre-training phase to reconstruct the masked depth image. (4) \textbf{3D policies}, including DP3~\cite{dp32024} and RVT-2~\cite{goyal2024rvt2learningprecisemanipulation}. \textit{\methodname} is compared with DP3 on MetaWorld and with RVT-2 on RLBench.

\subsection{Experiments on Simulation Benchmarks}

 \textbf{Evaluation Details.} 
Experiments in this section aim to evaluate various representation methods on simulation benchmarks for robotic manipulation (MetaWorld and RLBench).
% As shown in Fig.~\ref{fig:policy}, visual features extracted by each representation method are concatenated with robot state information and fed into an MLP policy head for action prediction. 
Each method is trained for 150 epochs, with 25 rollouts performed every 10 epochs, and the average success rate (SR) of the best performing model during training is reported. Note that our \textit{\methodname} focuses solely on the representation modules without modifying downstream robot action prediction models.

 \textbf{Results and Analysis.} 
As shown in Table~\ref{tab:main_experiment}, our \textit{\methodname} achieves average success rates of 82.6\% on MetaWorld and 82.0\% on RLBench, significantly outperforming all baselines.
The experimental results clearly reveal the superiority of 3D pre-training methods compared to their 2D counterparts. This indicates that even when 2D methods are pre-trained on larger-scale data, their performance is ultimately capped by an inherent lack of spatial awareness and multi-view ambiguity. For example, although VGGT demonstrates excellent performance in 3D reconstruction, its highly specialized training strategy impairs its generalization capabilities for robotics tasks that require broad scene understanding.
In contrast to these approaches, \textit{\methodname} acquires superior spatial awareness and geometric comprehension by pre-training directly on 3D data in a self-supervised manner within a unified spatial coordinate system, thereby achieving SOTA performance. Furthermore, \textit{\methodname} shows strong competitive performance against other advanced 3D policy methods like DP3\cite{dp32024} and RVT-2\cite{goyal2024rvt2learningprecisemanipulation}, proving that its extracted 3D spatial features are better aligned with the demands of robotic manipulation.

\begin{table}[]
\centering
\begin{center}
\caption{Quantitative Results (Success Rate, \%) on Policy Methods (RLBench)} 
\begin{tabular}{l|cccc|c}
\toprule
Method         & close b.   & close l. & water p. & toilet s. d. & Mean        \\ 
\midrule
PointNet \cite{pointnet2017}   & 52\%          & 88\%           & 20\%           & 96\%               & 64\%          \\
RVT-2(M) \cite{goyal2024rvt2learningprecisemanipulation}       & 88\%          & \textbf{100\%} & 12\%           & \textbf{100\%}     & 75\%          \\
RVT-2  \cite{goyal2024rvt2learningprecisemanipulation}    & \textbf{96\%} & 76\%           & 16\%           & 96\%               & 71\%          \\
Lift3D \cite{lift3d2024}    & 92\%          & 92\%           & 24\%           & \textbf{100\%}     & 77\%          \\
\midrule
\textbf{CLAR (Ours)}       & \textbf{96\%} & \textbf{100\%} & \textbf{32}\%           & \textbf{100\%}     & \textbf{82\%} \\ 
\bottomrule
\end{tabular}
\end{center}
\vspace{-7pt}
\flushleft
* close b., close l., water p., and toilet s. d. denote close box, close laptop lid, water plants, and toilet seat down, respectively. RVT-2(M) denotes RVT-2 in the multi-view setting.
    \vspace{-14pt}
\end{table}

\subsection{Real World Experiments}

 \textbf{Evaluation Details.} Our real-world evaluation is conducted on a 7-DOF Franka Emika robot arm, with a static side view captured by an Intel RealSense D455 RGB-D camera. In this setup, we compare our \textit{\methodname} against three baselines (VC-1, PointNet, Lift3D) on five tasks from RLBench~\cite{rlbench2020}: \textit{pick-banana}, \textit{put-bread}, \textit{water-flower}, \textit{place-cube-in-basin}, and \textit{open-drawer}. For each task, we collect 30 teleoperated demonstrations at 10Hz for training, with randomized initial states. The final trained policy is then evaluated 20 times per task. The policy outputs a 7D action space, comprising a 6D end-effector delta pose and a 1D binary gripper action. All experiments are run on an NVIDIA RTX A6000 GPU.

% \noindent  \textbf{Evaluation Details.}  This experiment is carried out to evaluate the performance of various methods in real-world scenarios. 
% We deploy three baselines (VC-1, PointNet, Lift3D) with our \textit{\methodname} and evaluate them on five tasks from RLBench~\cite{rlbench2020} (\textit{pick-banana}, \textit{put-bread}, \textit{water-flower},  \textit{place-cube-in-basin}, and \textit{open-drawer}). We collect 30 teleoperated demonstrations for training and test each task 20 times using the final epoch model. 
% We conduct all experiments using a 7-DOF Franka Emika robot arm, with a static side view captured by an Intel RealSense D455 RGB-D camera. The action space is 7-dimensional, including a 6-dimensional end-effector delta pose for the downstream impedance controller and a 1-dimensional binary gripper action. For each task, we collect 30 demonstrations at 10Hz via teleoperation, with a randomized initial state using either a scripted robot motion or manual resets by a human operator. Policies are performed using an NVIDIA RTX A6000 GPU.

\begin{figure}[ht]
% \captionsetup{skip=5pt}
    \centering
    \vspace{-2pt}
    \includegraphics[width=0.50\textwidth, ]{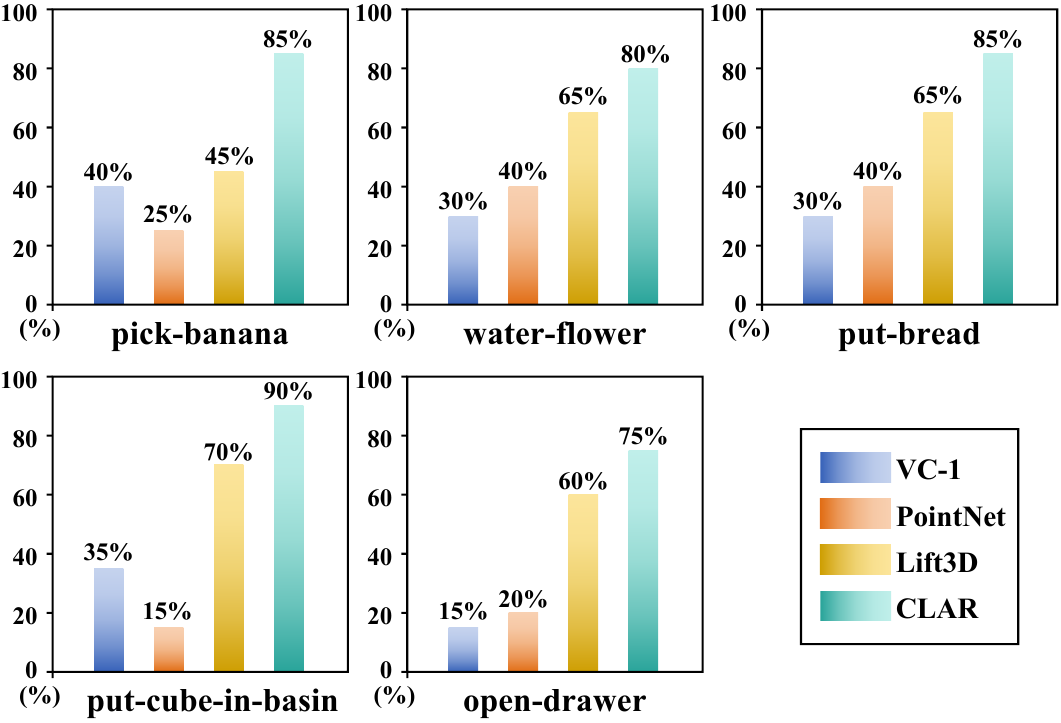}
    \vspace{-20pt}
    \caption{\textbf{Quantitative Results (Success Rate, \%) for Real World Experiments}.}
    \label{fig:real_world_result}
    \vspace{-7pt}
\end{figure}

\begin{figure*}[ht]
% \captionsetup{skip=5pt}
    \centering
    \includegraphics[width=0.95\textwidth, ]{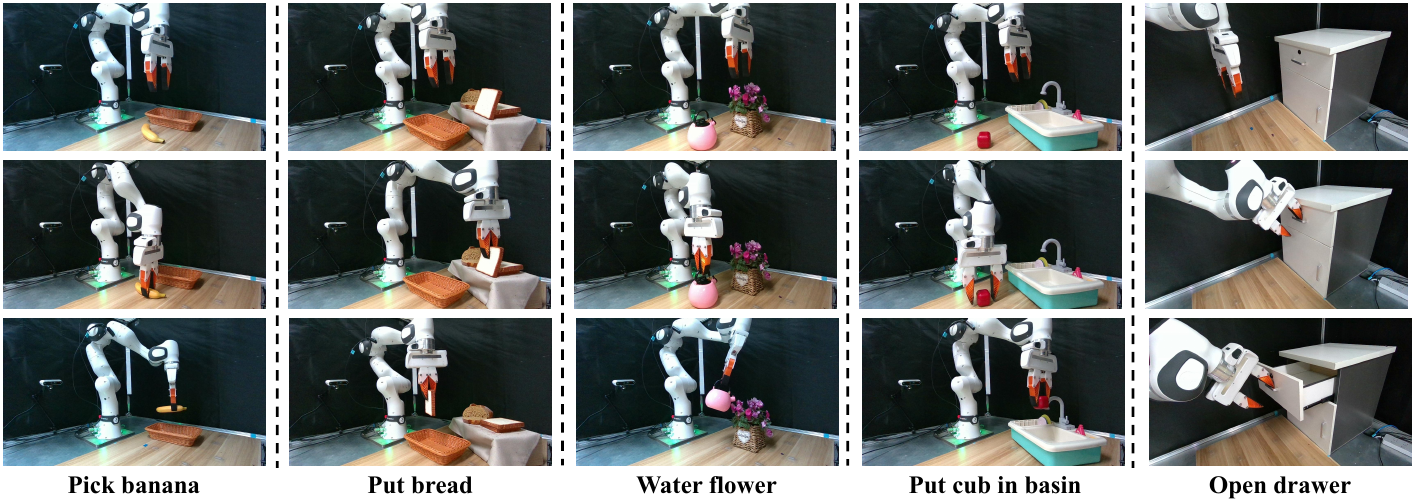}
    \vspace{-7pt}
    \caption{\textbf{Visualization of Real world Experiments}. The camera perspectives used in the experiments differ from the one used for visualization.}
    \label{fig:real_world}
    \vspace{-15pt}
\end{figure*}

% \noindent \textbf{Real World Setup.}
% For our real world setup, we conduct all experiments using a 7-DOF Franka Emika robot arm, with a static side view captured by an Intel RealSense D455 RGB-D camera. The action space is 7-dimensional, including a 6-dimensional end-effector delta pose for the downstream impedance controller and a 1-dimensional binary gripper action. For each task, we collect 30 demonstrations at 10Hz via teleoperation, with a randomized initial state using either a scripted robot motion or manual resets by a human operator. Policies are performed using an NVIDIA RTX A6000 GPU.

 \textbf{Results and Analysis.} As shown in Fig.~\ref{fig:real_world_result}, \textit{\methodname} achieves SOTA performance across multiple real-world tasks, demonstrating the robustness of its representation module for robotic manipulation.
The failures of the baseline methods highlight the advantages of our approach. The 2D-based method, VC-1, struggles to accurately perceive the geometry of fine structures, such as the kettle handle in the \textit{water-flower} task. The non-pre-trained PointNet performs poorly across all tasks due to its inability to adapt from limited few-shot demonstration data (30 trajectories). Even the 3D-aware Lift3D falters in tasks requiring precise manipulation (e.g., grasping in \textit{water-flower} or insertion in \textit{open-drawer}). This limitation stems from a modality mismatch, as it is pre-trained on 2D depth reconstruction but uses point clouds for downstream tasks, diminishing its pre-training effectiveness.
In contrast, \textit{\methodname} overcomes these issues by pre-training directly in 3D space with a unified coordinate system, leveraging both MAE and contrastive learning to enhance spatial and semantic understanding. This allows it to better comprehend object geometries and spatial relations, leading to superior real-world success rates.

\subsection{Multi-Task Experiments}

 \textbf{Evaluation Details.} Experiments in this section demonstrate the semantic understanding capabilities of different methods by introducing language instructions to enable robots to learn multi-task policies. We select three tasks from MetaWorld~\cite{metaworld2020}: \textit{button-press}, \textit{reach}, and \textit{lever-pull}. In addition, we generate several semantically similar robot instructions using ChatGPT for each task, such as "Move to the goal position," "Go to the target position," etc., resulting in a total of 18 instructions across the three tasks. Each method uses CLIP to extract text features, which are concatenated with visual features and robot state information before being fed into the policy head.
We employ multi-task training but evaluate each task separately to validate semantic understanding.

% \vspace{-12pt}
\begin{table}[ht]
\centering
% \caption{\textbf{Quantitative Results (Success Rate, \%) for Multi-task Experiments on MetaWorld} }
\caption{Quantitative Results (Success Rate, \%) for Multi-task Experiments on MetaWorld}
\scriptsize
\vspace{-5pt}
\begin{center}
\setlength{\tabcolsep}{3pt} % Adjust the space between columns
\renewcommand{\arraystretch}{1} % Adjust the row height (1 is the default)
\begin{tabular}{l|ccc|c}
\toprule
Method     & button-press & reach & lever-pull & Mean \\ 
\midrule
CLIP \cite{clip2021}       & 60\%           & 20\%    & 20\%         & 33.3\%   \\
R3M \cite{r3m2023}        & 40\%           & \textbf{60}\%    & 60\%         & 53.3\%   \\
VC-1 \cite{vc12023}       & 60\%           & \textbf{60}\%    & 80\%         & 66.6\%   \\
PointNet \cite{pointnet2017}   & 60\%           & 40\%    & 60\%         & 53.3\%   \\
PointNet++ \cite{pointnet++2017} & \textbf{80}\%           & 20\%    & 20\%         & 40.0\%   \\
PointNext \cite{pointnext2022}  & 60\%           & 40\%    & 20\%         & 40.0\%   \\
SPA \cite{spa2024}        & \textbf{80}\%           & 0\%     & \textbf{100}\%        & 60.0\%   \\
Lift3D \cite{lift3d2024}    & 60\%           & 40\%    & 80\%        & 60.0\%   \\ \midrule
\textbf{CLAR (Ours)}       & 60\%           & \textbf{60}\%    & \textbf{100}\%         & \textbf{73.3\%}   \\ 

\bottomrule
\end{tabular}
\end{center}
% \vspace{-15pt}
% \vspace{-8pt}
\label{tab:multitask}
    \vspace{-12pt}
\end{table}

 \textbf{Results and Analysis.} 
The multi-task experiment results in Table \ref{tab:multitask} show that \textit{\methodname} achieves an average success rate of 73.3\%. Models pre-trained on robotic datasets, such as Lift3D, SPA, VC-1, and R3M, generally surpass non-pre-trained 3D methods like PointNet, PointNet++, and PointNext. This suggests that these models benefit from the robust semantic understanding of pre-trained 2D foundation models. In contrast, CLIP, despite its strong general semantic understanding, performs poorly in robotic tasks due to the lack of pre-training in robotic scenarios. Our \textit{\methodname}, however, effectively transfers the semantic understanding of 2D foundation models to 3D through contrastive learning loss, ensuring alignment between the 3D feature space and the 2D foundation model's feature space. 
Finally, leveraging the semantic understanding distilled from 2D pre-trained model, \textit{\methodname} outperforms others and achieves a high success rate in this instruction-guided, multi-task imitation learning setting.

\subsection{Perspective Variation Experiments}

 \textbf{Evaluation Details.} This experiment examines whether perspective variations cause ambiguity in 2D representation methods and assesses their impact on 3D representation methods with a unified spatial coordinate system.
We evaluate our method against 2D representation methods (including R3M, VC-1, CLIP and Lift3D) on MetaWorld~\cite{metaworld2020}, using training data from the ``corner" perspective and testing under the ``corner2" perspective. Other experimental conditions and details remain consistent with the benchmark experiments.

% \vspace{-8pt}
\begin{table}[ht]
\centering
\caption{Quantitative Results (Success Rate, \%) for Perspective Variation Experiments on MetaWorld}
\scriptsize
\vspace{-8pt}
% \vspace{-15pt}
\begin{center}
\setlength{\tabcolsep}{3pt} % Adjust the space between columns
\renewcommand{\arraystretch}{1} % Adjust the row height (1 is the default)
\begin{tabular}{l|ccc|cc }
\toprule
Methods & button-press  & reach  & box-close & Mean \\
\midrule

% SGM~\cite{hirschmuller2008sgm} & 5.03 & 1.411 & 0.564 & 0.821 & 22.83 & 25.97 & 29.16 \\
% \midrule
CLIP \cite{clip2021} & 50\% (100\%) & 28\% (60\%) & 4\% (64\%) & 27.3\% (74.6\%)\\
R3M \cite{r3m2023}& 0\% (92\%) & 8\% (66\%) & 4\% (96\%) & 18.6\% (84.6\%)\\
VC-1 \cite{vc12023} & 8\% (92\%) & 28\% (36\%) & 4\% (66\%) & 13.3\% (64.6\%)\\
Lift3D \cite{lift3d2024} & 100\% (100\%) & 80\% (84\%) & 56\% (92\%) & 78.6\% (92\%)\\
\midrule
\textbf{CLAR (Ours)}  & \textbf{100\%} (100\%) & \textbf{80\%} (88\%) & \textbf{72\%} (96\%) & \textbf{84\%} (94.6\%)\\

\bottomrule
\end{tabular}
\end{center}
\vspace{-3pt}
% \vspace{-12pt}
\label{tab:perspective}
* $(\cdot)$ represents S.R. from the normal camera perspective.
    \vspace{-15pt}
\end{table}

 \textbf{Results and Analysis.} 
As shown in Table~\ref{tab:perspective}, 2D representation methods suffer a sharp performance drop under perspective variations, whereas 3D-based methods demonstrate significantly greater robustness. The core reason is that 2D representations lack a unified spatial coordinate system, making their understanding of object relations viewpoint-dependent. Even advanced methods like VC-1 and R3M, pre-trained on multi-view data, are confined to a camera-centric system and thus fail when the test viewpoint differs from the training one.
In contrast, 3D methods learn viewpoint-invariant spatial relationships by projecting point clouds into a unified coordinate system. This explains why both \textit{\methodname} and Lift3D exhibited slight performance degradation. Notably, the distinction between them is \textit{\methodname} unifies the spatial coordinate system during pre-training, whereas Lift3D's pre-training is camera-perspective-based. This fundamental difference allows our method to better handle perspective shifts, evidenced by a mere 10.6\% drop in its success rate.

% As shown in Table~\ref{tab:perspective}, 2D representation methods show a significant performance drop under perspective variation.
% This rapid decline results from the absence of a unified spatial coordinate system in 2D modality, causing ambiguity in spatial relative positions among objects in 2D-based methods across varying viewpoints. 
% Although VC-1 and R3M improve spatial perception through pre-training on multi-view and video data, their understanding remains limited to the camera coordinate system.
% Once there is a disparity between testing and training perspectives, 2D models inaccurately estimate object positions, leading to significant challenges in experiments with changing viewpoints, resulting in a notable decrease in success rates.
% Unlike 2D-based methods, 3D-based methods project point clouds into a unified coordinate system, enabling consistent learning of precise object positions regardless of viewpoint changes.  
% Both \textit{\methodname} and Lift3D demonstrated only slight performance degradation in this experiment, indicating an inherent robustness of 3D representations to viewpoint changes. Furthermore, since \textit{\methodname} unifies the spatial coordinate system of the point cloud during pre-training, whereas Lift3D is pre-trained solely from the camera's perspective, our method better handles such perspective variation case, resulting in a mere 10.6\% decrease in success rate.

\subsection{Ablation Study}

 \textbf{Evaluation Details.}
To validate the effectiveness of each module, we perform ablation experiments on \textit{\methodname} across four tasks from MetaWorld~\cite{metaworld2020}: \textit{hammer}, \textit{bin-picking}, \textit{assembly}, and \textit{shelf-place}. By removing the MAE loss, contrastive learning loss and local alignment loss separately, we examine the impact of each component on the robotic manipulation task. The average success rate measures the effect of each module.

\begin{table}[]
\scriptsize
\centering
\caption{Quantitative results (Success Rate, \%) for ablation study on MetaWorld} 
\vspace{-5pt}
\begin{tabular}{l|cccc|c}
\toprule
Method         & Pretrained & MAE & Con. & Local & Mean \\ \midrule
CLAR          & \ding{51}          & \ding{51}    & \ding{51}   & \ding{51}   & \textbf{83\%}    \\ 
CLAR  w/o MAE  & \ding{51}           & \ding{56}   & \ding{51}   & \ding{51}   & 73\%    \\
% CL3R  w/o local  & \ding{51}           & \ding{56}   & \ding{51}     & 53\%    \\
CLAR  w/o Local  & \ding{51}           & \ding{51}   & \ding{51}   & \ding{56}   & 77\%    \\
CLAR  w/o Local and Con.  & \ding{51}           & \ding{51}    & \ding{56}   & \ding{56}  & 70\%    \\
CLAR  w/o Pre. & \ding{56}           & \ding{56}   & \ding{56}  & \ding{56}  & 65\%    \\ \bottomrule
\end{tabular}
\vspace{-2pt}
% \vspace{-15pt}
\label{tab:ablation}
\flushleft
* Pretrained denotes whether the model has undergone pretraining, MAE refers to the use of a point cloud masked autoencoder, Con. indicates the application of contrastive loss, and Local signifies the exclusion of the local alignment loss.
\vspace{-15pt}
\end{table}

\begin{figure}[ht]
% \captionsetup{skip=5pt}
    \centering
    % \vspace{-10pt}
    \includegraphics[width=0.43\textwidth, ]{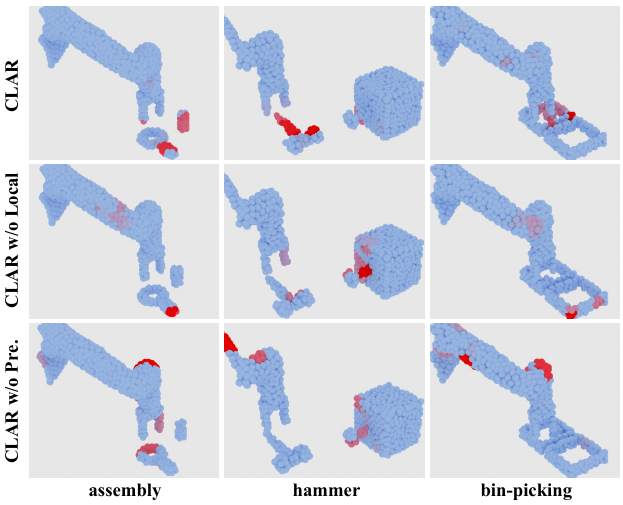}
    \vspace{-10pt}
    \caption{\textbf{Visualization of attention maps.}  Red highlights the regions with the highest attention scores.}
    \label{fig:atten}
    \vspace{-20pt}
\end{figure}

 \textbf{Results and Analysis.} From Table~\ref{tab:ablation}, we observe that both MAE and contrastive losses enhance the visual representation capabilities of the 3D pre-training model for robotic manipulation tasks during pre-training. Notably, MAE has the most significant impact in these tasks, as robust spatial awareness is more critical than semantic understanding when policy is required to tackle only one specific task. However, semantic understanding becomes equally important when the policy is required to tackle multiple tasks simultaneously. The results also show that incorporating contrastive learning alongside MAE does not hinder spatial understanding; instead, MAE loss and contrastive loss complement each other, further enhancing the 3D pre-training model's representation capabilities.
Furthermore, the local feature alignment module yields an 6\% gain for \textit{\methodname}. Visualizations of the attention maps (Fig.~\ref{fig:atten}) confirm that this component directs the model's focus toward task-relevant local geometric features, which in turn boosts its downstream imitation learning performance.

\section{CONCLUSION AND LIMITATIONS}

% We introduce \textit{\methodname}, a novel 3D pre-training framework designed to enhance the perception and representation learning for robotic manipulation policies, integrating powerful spatial awareness and extensive semantic understanding. Our framework also mitigates camera view ambiguity across diverse robotic datasets for pre-training and improves camera view generalizability. 
% Extensive experiments in both simulation and real-world demonstrate the superior performance of our framework in enabling powerful and robust visual representation in visuomotor robotic manipulation policy from various aspects.

In this work, we presented \textit{\methodname}, a novel 3D pre-training framework for robotic manipulation. Our approach bridges the gap between spatial-geometric and semantic representation learning by synergistically integrating a Masked Autoencoder with cross-modal contrastive learning. To make this fusion effective for manipulation, we introduced an adaptive local alignment mechanism to focus on fine-grained features, overcoming the limitations of conventional global alignment. Extensive experiments validate that \textit{\methodname} establishes a new state-of-the-art, providing a robust and generalizable representation for visuomotor policies.

% In this paper, we introduced \textit{\methodname}, a novel 3D pre-training framework designed to address the critical limitations of existing perception modules for robotic manipulation. Our work presents two primary innovations. First, we synergistically integrate two distinct pre-training paradigms: leveraging a Masked Autoencoder to build a robust understanding of 3D spatial geometry, while simultaneously using cross-modal contrastive learning to distill rich semantic knowledge from 2D foundation models. Our second key contribution, which makes this fusion effective for the nuanced demands of manipulation, is a novel adaptive local alignment mechanism. This mechanism forces fine-grained correspondence between 3D point cloud patches and 2D image features, ensuring that crucial local details are preserved. Extensive experiments in both simulation and the real world validate our approach, demonstrating that \textit{\methodname} achieves state-of-the-art performance and provides a more robust and generalizable representation for visuomotor policy learning.

\textbf{Limitations.}
\noindent A primary limitation is the scarcity of suitable 3D robotics datasets with precise camera and depth information, which restricts our pre-training scale compared to 2D foundation models. Furthermore, our experiments are confined to tabletop manipulation; while the model excels in this domain, its performance in more complex environments involving mobile navigation remains to be explored.

% \textbf{Limitations.}
% % One limitation of our current framework is the relatively coarse-grained semantic alignment with 2D foundation models. Currently, we align the point cloud feature only with the overall text feature derived from an entire sentence, which may overlook finer semantic details within the scene. In future work, we aim to explore more fine-grained alignment mechanisms that can associate point cloud features with localized semantic elements in the scene, which will further refine semantic understanding and improve the perception module’s ability to capture detailed contextual information.
% One limitation of our current framework is the relatively coarse-grained semantic alignment with 2D foundation models. Currently, we align the point cloud feature only with the overall text feature derived from an entire sentence, which may overlook finer semantic details within the scene. In future work, we aim to explore more fine-grained alignment mechanisms that can associate point cloud features with localized semantic elements in the scene, which will further refine semantic understanding and improve the perception module’s ability to capture detailed contextual information.

{
\bibliographystyle{IEEEtran}
\bibliography{main}
}

% Biography text here without a photo.

% \begin{IEEEbiography}[{\includegraphics[width=1in,height=1.25in,clip,keepaspectratio]{fig1.png}}]{IEEE Publications Technology Team}
% In this paragraph you can place your educational, professional background and research and other interests.\end{IEEEbiography}

\end{document}